\documentclass[11pt]{article}

\usepackage[preprint]{acl}

\usepackage{times}
\usepackage{latexsym}
\usepackage[T1]{fontenc}
\usepackage[utf8]{inputenc}
\usepackage{microtype}
\usepackage{inconsolata}
\usepackage{graphicx}

\usepackage{booktabs}
\usepackage{enumitem}
\usepackage{amssymb}
\usepackage{amsmath}
\usepackage{xcolor}

\title{ProACT: Towards Breakdown-Aware Proactive Agent \\ in Multi-User Collaboration}

\author{ 
    \textbf{Shu Yang\textsuperscript{1}},
    \textbf{Difei Xu\textsuperscript{1}},
    \textbf{Jiaxin Pei\textsuperscript{2}}, 
    \textbf{Di Wang\textsuperscript{1}†} \\[4pt]
    \textsuperscript{1}PRADA Lab, King Abdullah University of Science and Technology \\
    \textsuperscript{2}Stanford University
}

\begin{document}
\maketitle
\def\thefootnote{†}\footnotetext{Corresponding Author}
\begin{abstract}
Conversational agents are increasingly embedded in human collaborative work, yet they remain fundamentally \textit{passive} and \textit{reactive}: they respond to explicit user requests rather than proactively recognizing moments when a team would benefit from timely intervention as human collaborators often do. This reactive design substantially limits the use of agents as active participants in multi-user collaboration, where disagreements, ambiguous goals, forgotten constraints, underspecified plans, discussion loops, and imbalanced participation can gradually undermine group progress. 
To move agents from passive assistants toward active participants in multi-user collaboration, we introduce ProACT, a breakdown-aware agent framework grounded in theories of common ground, collaborative planning, and coordination work. 
ProACT observes the speaker-attributed conversation history, determines whether the current turn contains a collaboration breakdown requiring intervention, decides whether the agent should stay silent or speak, and, when speaking is needed, routes the case to a targeted collaboration skill.
We further introduce the first multi-user collaboration benchmark for evaluating proactive agents across project planning, product design, research collaboration, logistics, education, and resource-constrained decision making. Across 3,244 turn-level examples and five LLM backbones, ProACT consistently improves collaborative appropriateness, non-interruptiveness, conciseness, and judged intervention quality over direct chat. 

\end{abstract}

\section{Introduction}

Large language model (LLM) based agents are increasingly embedded in our collaborative workflows: they can help summarize meetings~\cite{kirstein2024tell}, draft plans~\cite{wei-etal-2025-plangenllms}, write code~\cite{jimenez2024swe}, and deliberations~\cite{abdelnabi2024negotiation,zhu2025can}. Despite this growing role, existing deployed agents remain fundamentally \textit{passive} and \textit{reactive}. They wait until the user explicitly asks a question, issues an instruction, or provides a predefined plan and prompts them to implement. This design misses many moments where an agent could actively participate in collaboration rather than merely assist on demand. 

\begin{figure}[h]
    \centering
\includegraphics[width=1\linewidth]{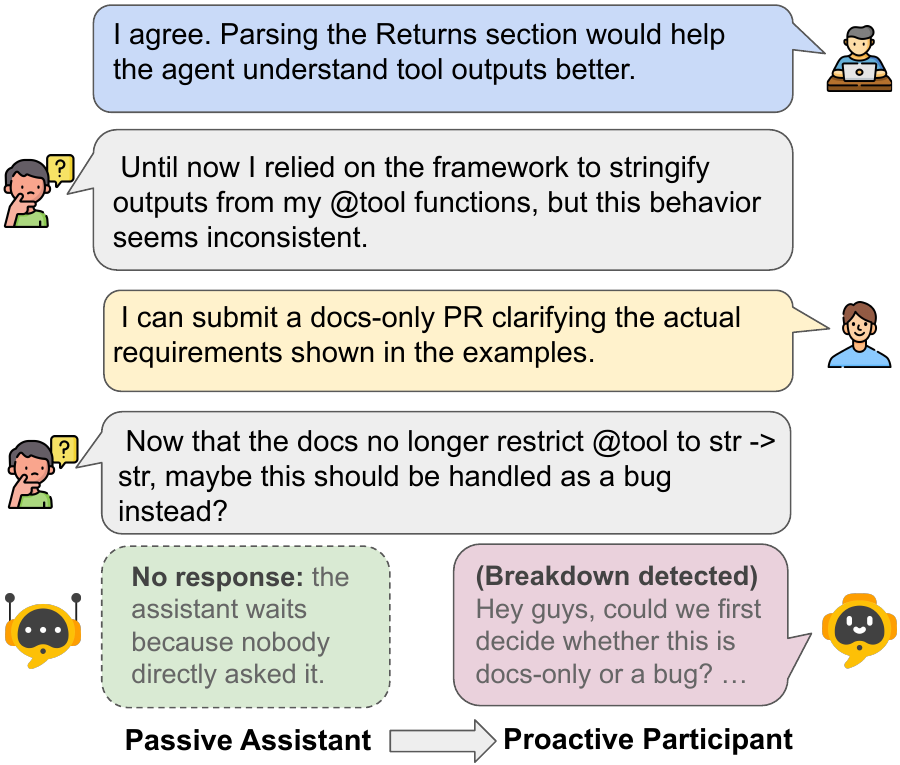}
    \caption{Illustration of moving agents from passive assistance to proactive participation in multi-user collaboration.}
    \label{fig:ProACTexample}
\end{figure}

Specifically, in multi-user scenarios, breakdowns often arise when participants lose alignment over shared understanding, commitments, responsibilities, or active constraints~\citep{clark1991grounding,grosz1996collaborative}. These failures are rarely abrupt; they accumulate through small coordination gaps. For example, one participant may optimize for speed while another optimizes for quality, creating a task or priority conflict \citep{Jehn1995AME}; a team may repeatedly revisit the same unresolved decision without recognizing that the discussion has entered a loop. 

In such cases, moving agents from passive
Assistance to proactive participation in multi-user collaboration can be quite helpful, as illustrated in Figure~\ref{fig:ProACTexample}. To achieve this, we propose ProACT, a breakdown-aware proactive framework for agents, which keeps observing the multi-party conversation and maintains a structured collaboration state. In contrast to prior work such as~\citet{lu2025proactive}, which focuses on predicting user intent from single-user input, we argue that proactive agents for collaboration require a different formulation from single-user proactive assistance. Rather than asking only how to help an individual user, a collaborative proactive agent must ask: \emph{When should I intervene in a group interaction where users may have different, uncertain, or conflicting intents?} This introduces social and normative challenges that are largely absent from single-user interaction. So ProACT continues to activately detect emerging breakdowns, such as conflict, uncertainty, underspecified plans, forgotten constraints, and discussion loops. When a breakdown requires intervention, 
ProACT selects a targeted collaboration skill and speaks only when a concise, neutral intervention is likely to improve coordination without interrupting productive discussion.

To validate ProACT, we introduce a benchmark for evaluating proactive agents in multi-user collaboration. Each example presents a multi-party conversation history and asks the agent to decide whether to remain silent or produce a concise group-facing intervention. The benchmark spans research collaboration, GitHub issue discussions, project coordination, product design, logistics, education, resource-constrained decision making, and other collaborative settings, yielding 3,244 turn-level examples. We evaluate agents from four perspectives: whether the intervention addresses an evidence-grounded breakdown, whether it avoids interrupting productive collaboration, whether it remains concise, and its overall intervention quality. Across five LLM backbones, ProACT consistently improves collaborative appropriateness, non-interruptiveness, conciseness, and judged intervention quality over baseline, showing that breakdown-aware skill routing helps agents decide both when to speak and how to intervene.

\section{Related Work}

\paragraph{Multi-user Collaboration}
\label{rw:multicollab}
Multi-user collaboration centers on joint social activity where participants maintain common ground, negotiate shared commitments, and coordinate interdependent actions toward shared or evolving goals~\citep{Bratman1992SharedCooperativeActivity,clark1991grounding,grosz1996collaborative,Malone1994Interdisciplinary}. In this setting, progress depends on whether participants remain aligned about what has been understood, what has been decided, who is responsible, which constraints remain active, and whose perspective is missing. Prior work shows that collaboration can be weakened by task and process conflict~\citep{Jehn1995AME}, failures of grounding and conversational repair~\citep{Schegloff1977Preference,clark1991grounding}, unmanaged dependencies and articulation work~\citep{Malone1994Interdisciplinary,Schmidt1992Taking}, inefficient group decision processes~\citep{DeSanctis1987Foundation,Briggs2003Collaboration}, and unequal participation or missing information in collective problem solving~\citep{Stasser1985Pooling,Woolley2010Evidence}. These studies suggest that effective collaboration requires continuous attention to shared understanding, responsibility, constraints, decision progress, and participation balance. This motivates proactive agents that can recognize emerging coordination problems and intervene only when a concise, neutral contribution is likely to help the group move forward.

\paragraph{LLM Agent for Coordination} 
Recent work on LLM agents has shown that language models can be organized as agents that reason, act, communicate, and coordinate through explicit interaction protocols. 
Early agent frameworks tightly couple reasoning with tool use, alternating between planning steps and external actions (e.g., calling APIs or using search)~\citep{yao2023react}, while multi-agent systems such as AutoGen, CAMEL, and MetaGPT coordinate multiple LLM agents through conversation, role assignment, workflow decomposition, or dynamic agent grouping~\citep{wu2024autogen,li2023camel,hong2024metagpt}. Recent systems such as Magentic-One adopt orchestrator-style architectures, where a lead agent plans, tracks progress, and delegates subtasks to specialized agents~\citep{fourney2024magentic}. Other work evaluates whether LLM agents can coordinate in practice, including coordination games, Theory-of-Mind belief tracking, and social-psychology-oriented analyses of consensus and debate~\citep{agashe2023llmcoordination,li2023theoryofmind,zhang2024collaboration}. Despite this progress, these settings largely focus on single-user task assistance or coordination \emph{among agents}, typically under a specified task or workflow, rather than an agent participating in ongoing \textit{human} multi-user collaboration. Our work studies an LLM agent embedded in a multi-party human discussion, where the key challenge is deciding \emph{when} to intervene proactively and \emph{how} to help without disrupting the group.

\section{Definitions}
\label{sec:def}
\subsection{Multi-User Collaboration Environment}

A multi-user collaboration environment is a shared communication space in which a set of participants $\mathcal{U}=\{u_1,\ldots,u_n\}$ work toward one or more tasks under evolving goals, constraints, roles, and commitments. At timestep $t$, the observed interaction prefix is represented as a speaker-attributed message sequence: $H_t = (x_i)_{i=1}^{t}$.
For each turn $i \in \{1,\ldots,t\}$, the message is defined as $x_i = (s_i, c_i)$,
where $s_i \in \mathcal{U}$ denotes the speaker and $c_i$ denotes the message content. Compared with single-user assistance, this environment involves multiple participants who may differ in goals, preferences, knowledge, responsibilities, authority as discribed by ~\citet{yang2026multi}. The agent observes the shared interaction history and decides whether to remain silent or contribute to the same communication channel. This setting makes proactive collaboration a timing and coordination problem: the agent must decide when its contribution would help the group move forward and when staying silent would avoid interrupting productive discussion.

\subsection{Collaboration Breakdown}

Following prior work on human collaboration in \S~\ref{rw:multicollab}, we define a \emph{collaboration breakdown} as an evidence-grounded state where the group may lose progress, shared understanding, coordination quality, or fair participation without timely repair. Disagreement, uncertainty, or delay becomes a breakdown when it blocks collaboration by obscuring intent, hiding trade-offs, violating constraints, underspecifying plans, repeating unresolved issues, or excluding relevant perspectives.

Table~\ref{tab:breakdowns} summarizes the breakdown categories used in this work. The taxonomy operationalizes recurring failure modes studied in human collaboration: task and process conflict~\citep{Jehn1995AME}, grounding and conversational repair~\citep{Schegloff1977Preference,clark1991grounding}, coordination and articulation work~\citep{Schmidt1992Taking,Malone1994Interdisciplinary}, group decision processes~\citep{DeSanctis1987Foundation,Briggs2003Collaboration}, and participation imbalance or missing perspectives in collective problem solving~\citep{Stasser1985Pooling,Woolley2010Evidence}. We use these categories to decide whether the current turn contains an intervention-worthy collaboration issue, while the detailed definitions are given in Table~\ref{tab:breakdowns}.

\begin{table*}[t]
\centering
\small
\begin{tabular}{p{0.22\linewidth} p{0.70\linewidth}}
\toprule
\textbf{Breakdown} & \textbf{Definition} \\
\midrule
Conflict & Participants express incompatible goals, preferences, assumptions, constraints, or decision criteria. \\
Uncertainty & Participants lack shared clarity about requirements, evidence, intent, terminology, or what decision needs to be made. \\
Underspecified plan & The group has a general direction, but important execution details such as owner, deadline, scope, dependency, metric, or next step remain unclear. \\
Forgotten constraint & The discussion overlooks an earlier constraint, commitment, decision, policy, budget, deadline, or technical requirement. \\
Goal drift & The conversation moves away from the shared objective, unresolved issue, or current decision point. \\
Discussion loop & Participants repeat the same issue, trade-off, or argument without adding new information or moving toward a decision. \\
Participation imbalance & The conversation is dominated by one perspective, or an important participant, stakeholder, or source of expertise is missing from the discussion. \\
Risky commitment & The group moves toward a decision or commitment before important risks, dependencies, assumptions, tests, or fallback plans are resolved. \\
\bottomrule
\end{tabular}
\caption{Collaboration breakdowns observed in multi-user interactions.}
\label{tab:breakdowns}
\end{table*}

\subsection{proactive Collaboration}

proactive collaboration refers to an agent's ability to decide, at each turn of a multi-user conversation, whether to remain silent or provide a group-facing intervention before being explicitly asked. The goal is to support the group process when a contribution can repair an emerging collaboration breakdown, while avoiding unnecessary interruption when participants are already making progress. Given the conversation history $H_t$, a proactive agent chooses an action $a_t \in \{\varnothing, u_t\},$ 
where $\varnothing$ denotes silence and $u_t$ denotes a visible group-facing intervention. A non-silence action is appropriate only when the conversation contains evidence of an intervention-worthy breakdown, the selected intervention matches the collaboration issue, and the response is likely to help the group move forward without disrupting productive discussion.

This definition connects directly to our evaluation criteria in \S~\ref{sec:evaluation-protocol}. An intervention should be \emph{appropriate}, meaning that it addresses an evidence-grounded collaboration issue; \emph{non-interruptive}, meaning that the agent remains silent when the group is already progressing or self-repairing; and \emph{concise}, meaning that the response supports the collaboration without dominating the conversation.

\section{ProACT: Breakdown-Aware proactive Agent in Multi-User Collaboration}
\label{sec:ProACT}
\begin{figure*}
    \centering
    \includegraphics[width=\linewidth]{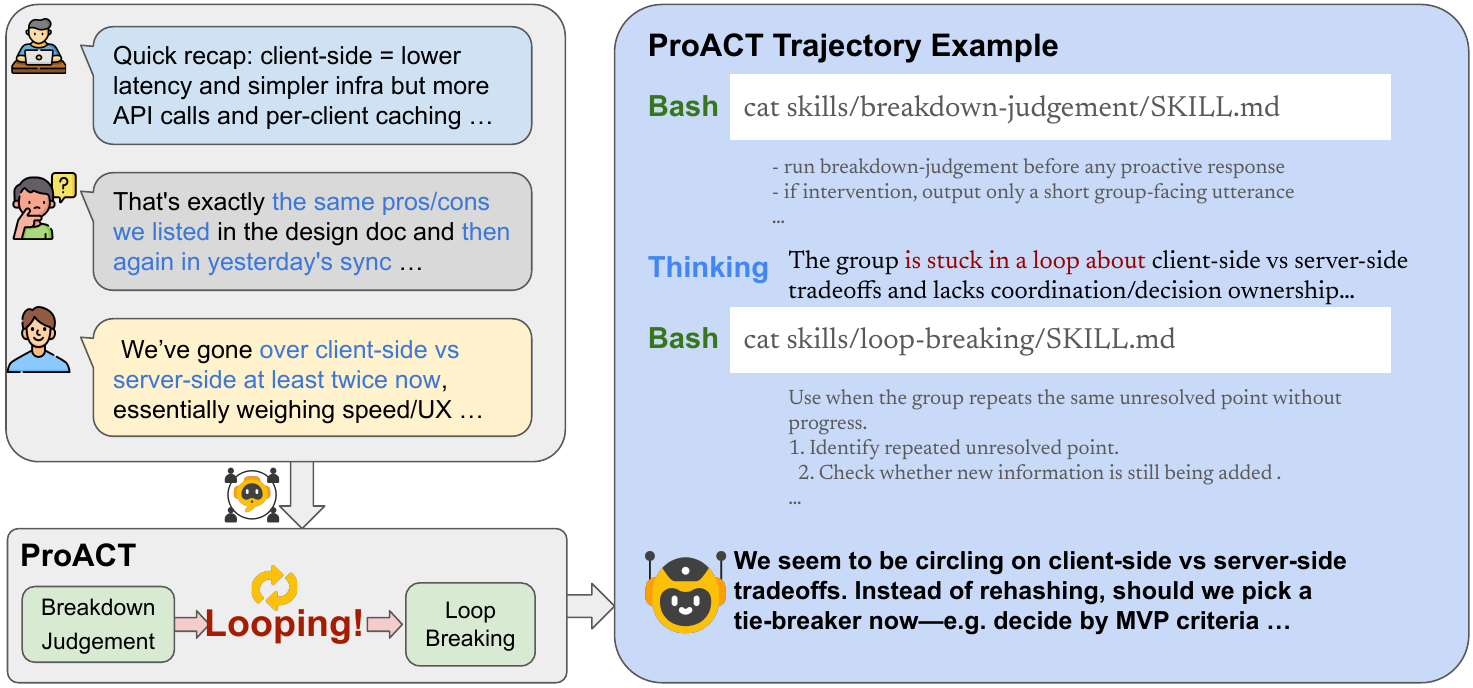}
    \caption{ProACT trajectory example. The agent monitors a multi-user conversation, detects emerging collaboration breakdowns such as discussion loops or conflicts, and applies targeted skills like loop-breaking or conflict mediation. The illustration shows how ProACT identifies a looping discussion and intervenes with a concise, evidence-grounded suggestion to guide the group toward resolution.}
    \label{fig:ProACToverview}
\end{figure*}

ProACT is a skill-guided framework for LLM-based proactive participation in multi-user collaboration. Building on the definitions in \S~\ref{sec:def}, ProACT observes the evolving speaker-attributed conversation history and decides whether the agent should remain \textit{silent} or produce a targeted group-facing \textit{intervention}. The framework first diagnoses whether the current turn contains an intervention-worthy collaboration breakdown, then routes intervention cases to a relevant collaboration skill. Each skill is specified as a lightweight operating procedure with a trigger condition, diagnostic cues, reasoning steps, social constraints, and an output contract. This structure maps proactive participation into an evidence-grounded decision process where interventions occur only when they are appropriate, non-interruptive, and concise. \S~\ref{sec:ProACT-pip} details the decision pipeline, \S~\ref{sec:ProACT-skills} presents the skill library, and \S~\ref{sec:ProACT-interface} describes the structured interface for routing evidence and executing skills.

\subsection{ProACT Decision Pipeline}
\label{sec:ProACT-pip}

ProACT maps an evolving multi-user conversation to either \textsc{Silence} or a targeted collaboration intervention. As illustrated in Figure~\ref{fig:ProACToverview}, the pipeline has three steps: diagnosing the current collaboration context, deciding whether the agent should speak, and routing intervention cases to an appropriate collaboration skill.

\paragraph{Breakdown-aware diagnosis.}
At turn \(t\), ProACT observes the conversation history \(H_t\), represented as speaker-attributed messages. The first step is to assess whether the current interaction contains an intervention-worthy collaboration breakdown:
\[
(b_t, z_t) = \mathrm{Diagnose}(H_t),
\]
where \(b_t\) denotes the detected collaboration issue, if any, and \(z_t \in \{0,1\}\) indicates whether an intervention is warranted. The diagnosis covers recurring breakdowns defined in Table~\ref{tab:breakdowns}. The decision also accounts for local context, such as who is speaking, whether a question is already directed to another participant, whether the group is actively making progress, and whether the evidence is strong enough for a concise intervention.

\paragraph{Skill-guided proactive action.}
If intervention is unwarranted, ProACT outputs \textsc{Silence}. If intervention is warranted, ProACT routes the detected issue to a relevant collaboration skill:
\[
k_t = r(b_t).
\]
The selected skill provides guidance for how the agent should help with that type of collaboration issue. It specifies the trigger condition, diagnostic cues, social constraints, and expected response format. The final action is:
\[
a_t = \pi_{k_t}(H_t), \qquad a_t \in \{\varnothing, u_t\},
\]
where \(\varnothing\) denotes silence and \(u_t\) denotes a visible group-facing intervention. When ProACT speaks, the response should be grounded in the conversation, address the current collaboration need, preserve human agency, and remain concise. In this way, ProACT treats proactive participation as a decision about when to help and how to help.

\subsection{Collaboration Skill Library}
\label{sec:ProACT-skills}

ProACT organizes proactive participation through a library of lightweight collaboration skills, summarized in Table~\ref{tab:skill-library}. In agent systems, a skill can be viewed as a reusable action routine: it specifies when the agent should use a capability, what evidence it should check, and how the final action should be formed. In our setting, these skills are designed for group-facing collaboration rather than tool execution. Each skill guides the agent toward a specific repair move, such as asking a clarification question, mediating a conflict, reminding the group of a prior constraint, or breaking a repeated discussion loop. Each skill is specified by four components: a trigger condition, diagnostic cues, social constraints, and an expected response format. The trigger condition describes when the skill may be useful. The diagnostic cues indicate what evidence should appear in the conversation history. The social constraints guide the agent to remain neutral, avoid unnecessary interruption, and preserve human agency. The response format keeps the final intervention concise and directly useful to the group. This design makes ProACT's interventions more controllable and interpretable: the model selects a skill based on the local conversation context, then produces a short group-facing utterance grounded in the detected collaboration need.

\subsection{Structured Skill Invocation Interface}
\label{sec:ProACT-interface}

ProACT uses a structured interface to separate internal skill selection from the visible group-facing response. Given the conversation history \(H_t\), the model first determines whether the current turn requires intervention. If no intervention is needed, it returns silence. If intervention is warranted, it identifies the collaboration issue, selects a relevant skill, and uses the skill instructions to produce the final action:
\[
H_t \rightarrow (b_t, k_t) \rightarrow a_t,
\qquad a_t \in \{\varnothing, u_t\},
\]
where \(b_t\) is the diagnosed collaboration issue, \(k_t\) is the selected skill, \(\varnothing\) denotes silence, and \(u_t\) is a visible group-facing utterance. The interface constrains the final output format. For silence, the visible response is empty. For intervention, the visible response contains only a concise message to the group, without diagnostic labels, skill names, rationale, or meta-commentary. This design keeps ProACT's runtime behavior simple and ensures that the evaluated output is exactly the agent's group-facing contribution.

\section{Evaluation for proactive Collaboration}
\label{sec:evaluation}

We construct a benchmark for evaluating proactive collaboration. Each example contains a multi-user conversation prefix \(H_t\) and a decision turn \(t\). The agent must choose between two actions: remain silent or produce a concise, group-facing intervention. This setup measures both timing and content: an effective agent should help when the conversation shows a collaboration breakdown and stay silent when human participants are already making progress.

\subsection{Dataset Construction}
\label{sec:dataset-construction}

We build the evaluation set from real multi-user conversations collected from public GitHub issue discussions and opensourced QMSum dataset~\citep{zhong2021qmsum}. These sources cover software collaboration, product design, research meetings, and group decision-making. We convert each conversation into a sequence of decision points, where each point asks whether a proactive agent should stay silent or intervene.

\begin{figure}[t]
\centering
\includegraphics[width=\columnwidth]{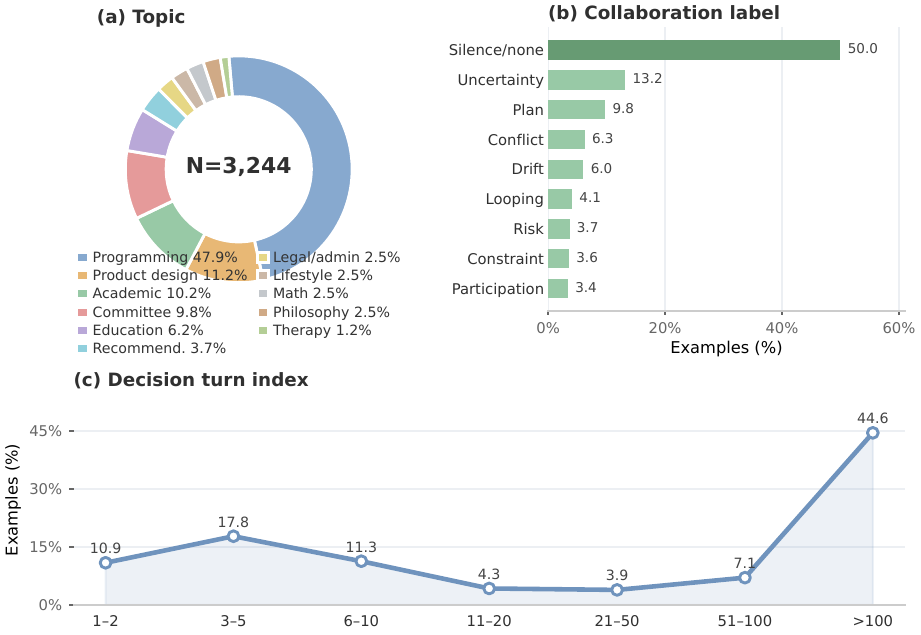}
\caption{Distribution of the proactive collaboration evaluation dataset by topic, collaboration label, and decision position. Percentages are computed over all 3,244 examples.}
\label{fig:dataset-distribution}
\end{figure}

We pre-label candidate decision points using multiple LLM annotators. For each point, annotators decide whether a proactive agent should remain silent or intervene. For intervention cases, they also identify the collaboration issue and provide a short rationale. We describe the annotation protocol in Appendix~\ref{app:llm-annotation}. Candidate points with annotator disagreement or ambiguous evidence are sent to human review, yielding 694 human-reviewed examples. After annotation, we construct a balanced evaluation set with equal numbers of intervention and silence cases. To improve coverage of rare collaboration failures, we add 800 synthetic conversations derived from BEAM~\citep{tavakoli2025beyond}, which have more interaction turns. We adapt BEAM scenarios into multi-user discussions and inject controlled breakdowns from our taxonomy, such as forgotten constraints, discussion loops, participation imbalance, and risky commitments. These synthetic examples supplement the real conversations; details of the construction procedure are provided in Appendix~\ref{app:synthetic-data}. The final dataset contains 3,244 examples: 2,444 real examples and 800 synthetic examples, balanced into 1,622 intervention cases and 1,622 silence cases. Figure~\ref{fig:dataset-distribution} shows the distribution by topic, collaboration label, and decision position.

\subsection{Evaluation Protocol}
\label{sec:evaluation-protocol}

We compare ProACT with a Direct Chat baseline under matched input conditions. For each example, both methods receive the same conversation prefix \(H_t\). Direct Chat generates the next assistant response directly, while ProACT uses breakdown diagnosis and skill guidance to decide whether to stay silent or produce a concise group-facing intervention. We evaluate outputs with an LLM. The judge evaluates four criteria. \textbf{Appropriate} measures whether the agent's action fits the collaboration state: intervening when there is an evidence-grounded breakdown and staying silent when the group is already making progress. \textbf{Non-interruptive} measures whether the output avoids disrupting productive human discussion, duplicating an ongoing repair, taking sides prematurely, or dominating the group. \textbf{Concise} measures whether the visible response is short, group-facing, and free of unnecessary explanation or meta-commentary. \textbf{Quality} is a 1--5 score for visible interventions, reflecting whether the response is useful, grounded in the conversation, neutral, and actionable. For each example \(i\), the judge returns binary indicators of appropriateness, non-interruption, and conciseness. We report Appropriate, Non-int., and Concise as the average of these indicators over the dataset \(\mathcal{D}\), which corresponds to the fraction of examples receiving a positive judgment for each criterion. For Quality, we report the average 1--5 judge score over visible interventions on reference-intervention examples, so response quality is measured separately from abstention behavior. Full judging details are provided in Appendix~\ref{app:judge-prompt}.

\section{Experiments}

\begin{figure*}[ht]
\centering
\includegraphics[width=\textwidth,trim={0 25pt 0 0},clip]{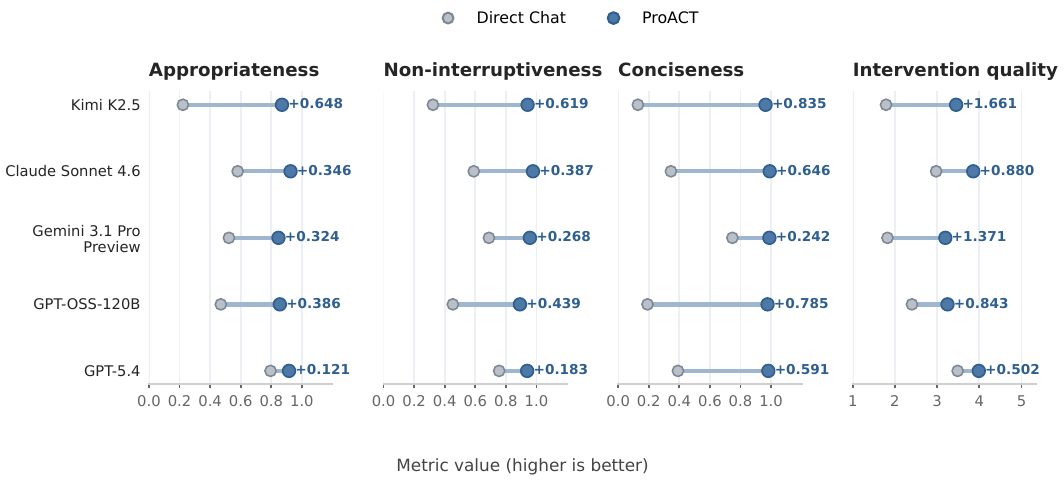}
\caption{proactive collaboration evaluation results across five LLM backbones. Grey markers show Direct Chat and blue markers show our ProACT.}
\label{fig:ProACT-main-results}
\end{figure*}

\subsection{Experimental Setup} 
We evaluate five agent backbones on the full 3,244-example test set: GPT-5.4, Kimi K2.5, Claude Sonnet 4.6, Gemini 3.1 Pro Preview, and GPT-OSS-120B. For each model, we compare two methods under matched input conditions: Direct Chat and ProACT. Direct Chat generates the next assistant message in a single pass. ProACT first diagnoses whether the current turn requires intervention, then either remains silent or invokes a collaboration skill to produce a concise group-facing intervention.
For all models, we use the same decoding configuration where supported: temperature \(=1.0\) and top-\(p=0.98\). All outputs are evaluated by GPT-5.4-Nano as an external judge following the protocol in \S~\ref{sec:evaluation-protocol}.

\subsection{Results}
\begin{figure*}[t]
\centering
\includegraphics[width=\textwidth]{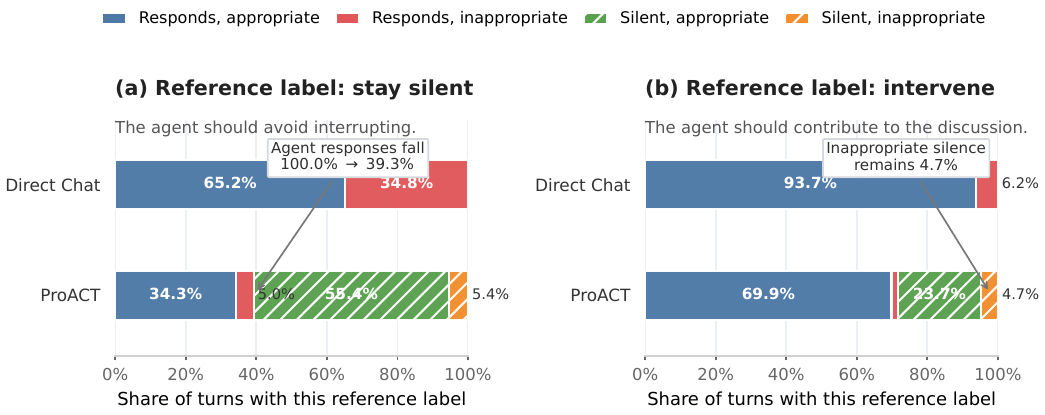}
\caption{GPT-5.4 participation timing by reference label. Each stacked bar decomposes the model behavior into whether the agent responds or stays silent. ProACT sharply reduces unnecessary responses when the reference label prefers silence.}
\label{fig:ProACT-when-to-speak}
\end{figure*}

\paragraph{ProACT consistently improves collaborative participation.}
Figure~\ref{fig:ProACT-main-results} compares ProACT with the Direct Chat baseline across five LLM backbones. ProACT improves every reported metric for every model, including collaborative appropriateness, non-interruptiveness, conciseness, and intervention quality. The gains are largest for Kimi K2.5: appropriateness increases from 0.222 to 0.870, non-interruptiveness from 0.323 to 0.942, conciseness from 0.129 to 0.964, and intervention quality from 1.789 to 3.450. The same trend holds for Claude Sonnet 4.6, Gemini 3.1 Pro Preview, GPT-OSS-120B, and GPT-5.4. These consistent improvements show that ProACT helps agents decide both when to participate and how to contribute, producing interventions that are more appropriate, less disruptive, more concise, and higher quality than direct chat responses.

\paragraph{ProACT improves participation timing.}
The gains in Figure~\ref{fig:ProACT-main-results} are partly driven by better participation timing. Figure~\ref{fig:ProACT-when-to-speak} decomposes GPT-5.4 behavior by reference label and judges appropriateness. On silence-labeled examples, Direct Chat responds on every turn, with 34.8\% of responses judged inappropriate. ProACT responds on only 39.3\% of these turns and stays silent in most remaining cases, increasing overall judged appropriateness from 65.2\% to 89.6\%. On intervention-labeled examples, ProACT responds less often than Direct Chat, but its visible interventions are of higher quality, improving from 3.486 to 3.988. Inappropriate silence remains low at 4.7\%. These results show that ProACT improves participation timing: it avoids unnecessary interruptions when the group is already progressing, while still producing useful interventions when collaboration needs support.

\begin{figure}[t]
\centering
\includegraphics[width=\columnwidth]{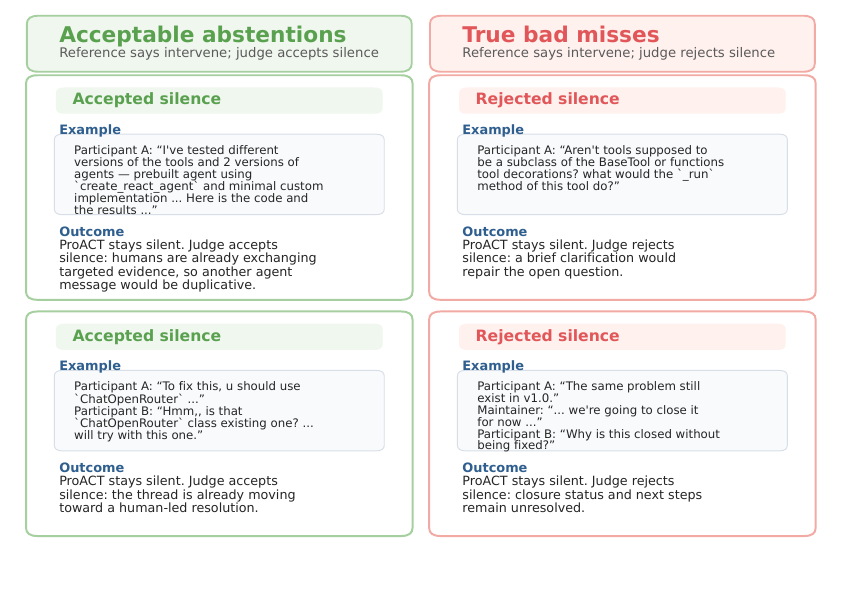}
\caption{Real examples of ProACT silence on intervention-labeled turns. Some apparent misses are acceptable abstentions because participants are already repairing the issue, while true bad misses involve clearer unresolved coordination needs.}
\label{fig:ProACT-abstention-examples}
\end{figure}

\begin{figure}[t]
\centering
\includegraphics[width=0.5\textwidth]{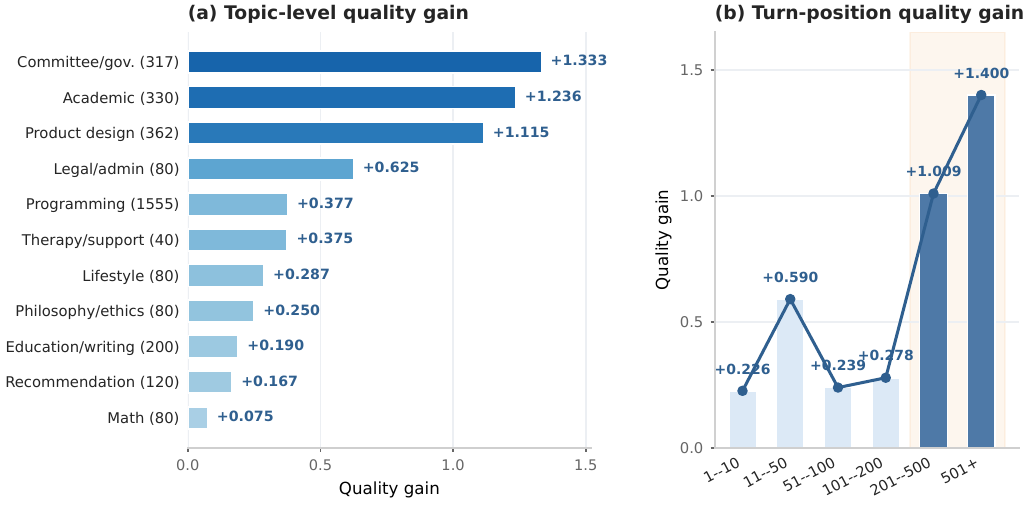}
\caption{GPT-5.4 intervention quality gains by topic and decision position. Values show the absolute improvement of ProACT over Direct Chat.}
\label{fig:ProACT-topic-turn-quality}
\end{figure}

\paragraph{Quality gains are strongest in social planning tasks and later turns.}
Figure~\ref{fig:ProACT-topic-turn-quality} analyzes where ProACT improves GPT-5.4 intervention quality. The largest gains appear in topics that require social coordination and planning. Committee and governance discussions improve by 1.333 quality points, academic collaboration by 1.236, and product design by 1.115. Programming also improves by 0.377, but the gain is smaller, likely because many programming examples already contain explicit technical questions, error traces, or reproduction details that make direct responses easier. The analysis by conversation position shows a similar pattern. ProACT improves intervention quality across the conversation, with larger gains in later stages. The quality gain is 0.226 points within the first 10 turns, 0.590 points for turns 11 to 50, 1.009 points for turns 201 to 500, and 1.400 points after turn 500. These results suggest that ProACT is especially useful when collaboration accumulates unresolved context, repeated decisions, or coordination failures, where skill-guided interventions provide more value than direct chat responses.

\paragraph{Most apparent misses reflect conservative abstention.}
A concern is that ProACT may improve non-interruptiveness by staying silent too often. We examine intervention-labeled examples where GPT-5.4 ProACT produces no response. Although these cases appear to be missed interventions under the reference label, most are judged acceptable: 83.5\% of ProACT's silence decisions in this subset are accepted by the external judge, and rejected silences account for only 4.7\% of all intervention-labeled examples.

The qualitative pattern in Figure~\ref{fig:ProACT-abstention-examples} explains this result. In accepted-silence cases, participants are already exchanging relevant evidence, answering each other, or moving toward a human-led resolution. An additional agent message would likely duplicate the ongoing repair or interrupt productive discussion. Rejected-silence cases are different: they contain clearer unresolved coordination needs, such as an unanswered clarification question or an unresolved challenge to issue closure. This suggests that many apparent misses reflect ambiguity in proactive timing, not simple failures to help. ProACT tends to be conservative about speaking, which improves group-facing behavior, while leaving room for stronger detection of cases that require clarification or common-ground repair.

\section{Conclusion}
We introduced ProACT, a breakdown-aware framework for proactive agents in multi-user collaboration. ProACT observes speaker-attributed conversation history, detects intervention-worthy collaboration breakdowns, and chooses between silence and a concise group-facing intervention through a targeted skill library. We also built a new benchmark from real multi-user conversations and BEAM-derived synthetic cases. Across five LLM backbones, ProACT improves appropriateness, non-interruptiveness, conciseness, and intervention quality over Direct Chat. Further analysis shows that these gains come from better participation timing: ProACT reduces unnecessary interruptions while preserving useful interventions when collaboration needs support. The strongest gains appear in social planning tasks and later-stage conversations, where accumulated context and unresolved coordination issues make direct responses less reliable. We hope this work serves as a starting point for building proactive agents that support human collaboration through timely participation.

\newpage
\section*{Limitations}
This work evaluates ProACT in a controlled offline, turn-level setting. This design enables consistent comparison because each model receives the same conversation history and must make the same silence-or-intervention decision, but it does not capture every dynamic of live collaboration, such as trust formation, user adaptation to repeated agent interventions, or downstream reactions after an intervention. We also use an external LLM judge to scale evaluation of appropriateness, non-interruption, conciseness, and quality, which supports broad comparison across models but should be complemented by future human studies. These limitations motivate future work on live multi-user deployments, richer cultural and organizational settings, and longer-horizon measures such as trust, perceived fairness, decision quality, coordination cost, and participants' ability to contest or repair agent behavior. 

Also, although ProACT is designed to support collaboration through concise and neutral interventions, proactive participation also introduces risks that are not fully captured. In real multi-user settings, an agent that decides when to speak may gradually change the mode of human collaboration: participants may defer to the agent, treat its interventions as authoritative, or rely on it to structure disagreement and decision making. More importantly, the same mechanism that helps repair coordination breakdowns could be misused or misaligned to steer group behavior. For example, a proactive agent might selectively surface evidence, frame options asymmetrically, suppress dissent, or induce participants to cooperate toward goals that they have not explicitly endorsed. Such risks relate to model deception, persuasion, and manipulation in group decision processes. Future work should therefore include controlled human-subject studies and deployment audits to examine how proactive agents affect trust, agency, participation balance, dissent, decision quality, and susceptibility to manipulation. Practical deployments should also include transparency, logging, contestability, and role or authority constraints so that users can recognize, challenge, or disable agent interventions.

\bibliography{references}

\newpage
\appendix
\definecolor{judgeboxborder}{RGB}{180,180,180}
\definecolor{judgeboxbg}{RGB}{248,248,248}
\definecolor{promptboxbg}{HTML}{F9FAFB}
\definecolor{promptboxborder}{HTML}{D1D5DB}
\definecolor{promptboxaccent}{HTML}{111827}

\clearpage
\section{Dataset Construction and Annotation Details}
\label{app:llm-annotation}
We construct the benchmark by converting multi-user conversations into decision-point examples for proactive collaboration. Given a source conversation \((x_1,\ldots,x_T)\), a candidate example at turn \(t\) contains the conversation prefix \(H_t=(x_i)_{i=1}^{t}\). The annotation task asks whether a proactive agent should remain silent or produce a short group-facing intervention at that point. Because proactive timing can be context-dependent, we treat the labels as reference annotations rather than absolute ground truth.

\paragraph{Source artifact provenance and redistribution.}
For each source used in benchmark construction, we retain provenance metadata, including the source type and source identifier when available. Public collaborative discussions are used as examples of naturally occurring multi-user coordination, while existing research datasets and scenarios are used only for research evaluation and controlled scenario construction. We do not treat access to an artifact as permission for unrestricted redistribution.

\begin{table*}[t]
\centering
\begingroup
\setlength{\fboxsep}{10pt}
\setlength{\fboxrule}{0.8pt}
\fcolorbox{judgeboxborder}{judgeboxbg}{%
\begin{minipage}{0.94\textwidth}
\small

\textbf{LLM annotation prompt template.}

\medskip
\textbf{System instruction.}
You are annotating a multi-user collaboration conversation for a proactive assistant. Given the conversation prefix up to the current decision point, decide whether the assistant should stay silent or intervene now.

\medskip
\textbf{Task.}
Return a structured annotation with three fields. 
\texttt{agent\_action} should be either \texttt{silence} or \texttt{intervene}. 
\texttt{breakdown\_type} should be one of \texttt{none}, \texttt{uncertainty}, \texttt{underspecified\_plan}, \texttt{conflict}, \texttt{forgotten\_constraint}, \texttt{drift}, \texttt{looping}, \texttt{imbalanced\_participation}, or \texttt{risky\_commitment}. 
\texttt{rationale} should be a brief explanation grounded in specific evidence from the visible conversation.

\medskip
\textbf{Decision criteria.}
Choose \texttt{silence} when participants are already making progress, when a participant has already asked the needed clarification, when the issue is being self-repaired, when the evidence is weak, or when an assistant message would be duplicative or interruptive. Choose \texttt{intervene} only when the conversation shows an unresolved collaboration breakdown and a short group-facing message would likely improve coordination.

\medskip
\textbf{Output format.}
Return JSON only:
\begin{quote}
\small\ttfamily
\{\\
\quad "agent\_action": "silence" or "intervene",\\
\quad "breakdown\_type": "...",\\
\quad "rationale": "..."\\
\}
\end{quote}

\end{minipage}%
}
\endgroup
\caption{LLM annotation prompt used for pre-labeling proactive collaboration decision-point examples.}
\label{tab:llm-annotation-prompt}
\end{table*}

\paragraph{Personally identifying information and offensive content.}
Because our real examples are derived from public collaborative discussions and existing benchmark conversations, they may contain public user handles, names, project-specific references, URLs, email addresses, or other strings that could identify individuals or organizations. We therefore screen collected examples before inclusion. We use automatic pattern checks for common identifiers, including email addresses, phone numbers, URLs, social-media or GitHub handles, and speaker-name fields, followed by manual inspection during candidate filtering and human adjudication. In the benchmark release and in paper examples, speaker names and user handles are replaced with role-neutral identifiers such as \texttt{Participant A}, \texttt{Participant B}, or \texttt{Maintainer}. Direct contact information, private links, and unnecessary project-specific identifiers are removed or masked. Reviewer identifiers used during annotation are used only for assignment and adjudication and are not included in the released data.
\paragraph{Candidate extraction.}
We begin with real multi-user conversations from public collaborative settings, including GitHub issue discussions and QMSum meetings. These sources cover software collaboration, product design, research meetings, and group decision-making. We segment each conversation into candidate decision points and keep cases where the prefix contains enough context for an agent to judge the collaboration state. We remove examples that are too short, lack multiple participants, contain insufficient context, or do not involve a collaborative process.

\paragraph{LLM pre-labeling.}
Each candidate decision point is first pre-labeled by multiple LLM annotators. The annotators receive the conversation prefix and produce three fields: (1) a decision label \(y_t \in \{\textsc{Silence}, \textsc{Intervention}\}\); (2) a collaboration issue label, selected from the breakdown categories in Table~\ref{tab:breakdowns} or \textsc{None}; and (3) a short rationale grounded in the visible conversation. Annotators are instructed to choose \textsc{Silence} when participants are already making progress, when the issue is being self-repaired, when evidence for a breakdown is weak, or when an agent response would mainly duplicate the ongoing discussion.

\paragraph{Human review and adjudication.}
LLM pre-labeling is used as a screening step, not as the final annotation. We identify 740 candidate examples with LLM annotator disagreement, unclear evidence, malformed rationales, or ambiguous intervention timing. Two CS Ph.D. students review this subset by inspecting the conversation prefix, the proposed decision label, the breakdown type, and the rationale. They verify whether intervention cases contain concrete evidence of a collaboration breakdown and whether silence cases reflect productive ongoing human collaboration. After adjudication, we retain 694 examples with aligned final labels and remove cases that remain underspecified, out of scope, or too ambiguous for reliable evaluation.
Reviewers were instructed to answer the following question: ``Given the collaboration history up to the current turn, should a proactive agent intervene now? If yes, what collaboration breakdown is happening, which skill should be used, and what should the agent say?'' The core rule was: set \texttt{should\_intervene=true} only when speaking now would likely improve the group process at the current turn. A collaboration breakdown alone was not sufficient; if the group was already resolving the issue, the agent should remain silent. For each candidate prefix, reviewers were asked to: (1) read the full prefix; (2) identify the current collaboration state; (3) ask whether a helpful facilitator would speak now; (4) if yes, choose the breakdown type and preferred skill; (5) mark evidence turns; (6) write a brief reference intervention; (7) optionally write a bad intervention; and (8) write an annotation rationale. Reviewers used \texttt{silence} when there was no breakdown, the issue was low-stakes or already resolved, the evidence was insufficient, or speaking would duplicate or interrupt human facilitation.

\begin{table*}[t]
\centering
\small
\begin{tabular}{p{0.22\textwidth}p{0.72\textwidth}}
\toprule
\textbf{UI region} & \textbf{Displayed content / editable fields} \\
\midrule
Header controls & Reviewer ID, subdataset filter, action filter, breakdown filter, review status filter, and search box. \\
Conversation panel & Recent conversation context, current turn highlighting, evidence-turn highlighting, ``Load full prefix'' button, and original model annotation. \\
Human annotation panel & Review status, \texttt{should\_intervene}, \texttt{agent\_action}, \texttt{breakdown\_type}, \texttt{preferred\_skill}, \texttt{evidence\_turns}, \texttt{missing\_plan\_slots}, \texttt{reference\_intervention}, \texttt{bad\_intervention}, \texttt{annotation\_rationale}, optional human comment, and save/autosave controls. \\
\bottomrule
\end{tabular}
\caption{Textual reconstruction of the human-review interface used for validating proactive-collaboration reference labels. No separate risk or consent disclaimer was displayed in the review UI; reviewers only labeled existing public/benchmark or synthetic text excerpts and could mark examples as \texttt{skip} or \texttt{needs\_discussion}.}
\label{tab:human-review-interface}
\end{table*}
\section{Synthetic Data Construction}
\label{app:synthetic-data}

We use synthetic examples to improve coverage of collaboration failures that are important for proactive agents but appear less frequently in public conversations. The synthetic data are derived from BEAM~\citep{tavakoli2025beyond}, which provides long-context task scenarios that can be adapted into multi-user collaborative settings. We use BEAM as a source of task contexts, constraints, and decision situations, then inject short multi-user episodes that create controlled collaboration breakdowns.

\paragraph{Scenario adaptation.}
For each selected BEAM context, we identify the task goal, relevant participants, available information, constraints, and possible decision points. We then adapt the context into a short group-chat episode in which multiple participants exchange information, propose options, and move toward a decision. The adapted scenarios cover project planning, product design, logistics, education, resource allocation, research collaboration, and other collaborative tasks.

\paragraph{Synthetic breakdown injection.}
Each synthetic example targets one breakdown category from Table~\ref{tab:breakdowns}. The injected episode contains setup or evidence turns followed by a final trigger turn. The setup establishes the relevant context, such as an earlier constraint, an unresolved trade-off, a missing stakeholder, or an incomplete plan. The final trigger turn makes the collaboration issue visible at the current decision point. This design ensures that the proactive intervention is useful at that moment, rather than too early or only later.

\paragraph{Generation prompt.}
Table~\ref{tab:synthetic-generation-prompt} shows the prompt template used to generate BEAM-derived injected breakdown episodes. The generator is instructed to produce only natural user/team messages. It must not reveal annotation labels, breakdown names, preferred skills, ground truth, or any reference to the proactive assistant. This prevents the synthetic conversation from containing artificial cues that would make the task easier than real collaboration.

\begin{table*}[t]
\centering
\begingroup
\setlength{\fboxsep}{10pt}
\setlength{\fboxrule}{0.8pt}
\fcolorbox{judgeboxborder}{judgeboxbg}{%
\begin{minipage}{0.94\textwidth}
\small

\textbf{Synthetic breakdown injection prompt template.}

\medskip
\textbf{System instruction.}
You generate realistic multi-user collaboration chat snippets for synthetic data. Return JSON only.

\medskip
\textbf{User instruction.}
Generate only the injected user chat turns for a synthetic multi-user collaboration dataset. A long conversation is used as background. Create a two-stage group-chat episode to insert into that conversation. The episode must contain natural user or team messages, not labels or analysis. Do not mention annotation, dataset construction, breakdown type, preferred skill, ground truth, labels, or the AI assistant in the injected turns.

\medskip
Return the setup or evidence messages first, then the final trigger message last. The setup messages will be inserted earlier in the conversation, and the final trigger message will become the latest visible message after which a proactive collaboration assistant should decide whether to intervene. Use 3--6 injected turns from at least three human collaborators. Keep messages concise and realistic.

\medskip
\textbf{Target breakdown:} \texttt{\{breakdown\_type\}}

\textbf{Definition:} \texttt{\{breakdown\_definition\}}

\medskip
\textbf{Ground-truth requirements.}
The group must not resolve the issue before the final injected turn. The final injected turn must connect back to the earlier setup or evidence. The final injected turn must make the intervention useful at the current decision point. Avoid cues for other labels and instantiate the specified breakdown rather than generic confusion.

\medskip
\textbf{Positive pattern:} \texttt{\{positive\_pattern\}}

\textbf{Negative pattern to avoid:} \texttt{\{negative\_pattern\}}

\textbf{Recent BEAM context before insertion:} \texttt{\{recent\_beam\_context\}}

\medskip
\textbf{Output format.}
Return only a JSON array:
\begin{quote}
\small\ttfamily
[\\
\quad \{ "speaker": "participant name", "text": "message text" \},\\
\quad ...\\
]
\end{quote}

\end{minipage}%
}
\endgroup
\caption{Prompt template used to generate synthetic BEAM-derived collaboration breakdown episodes.}
\label{tab:synthetic-generation-prompt}
\end{table*}

\paragraph{Breakdown-specific constraints.}
Table~\ref{tab:synthetic-breakdown-guide} summarizes the generation constraints used for each breakdown type. These constraints are designed to create examples where the target breakdown is visible from the conversation prefix while avoiding confounds with other labels.

\begin{table*}[t]
\centering
\small
\setlength{\tabcolsep}{5pt}
\begin{tabular}{p{0.19\textwidth}p{0.36\textwidth}p{0.36\textwidth}}
\toprule
\textbf{Breakdown} & \textbf{Generation target} & \textbf{Pattern to avoid} \\
\midrule
\textsc{Conflict} &
Two participants hold incompatible goals, constraints, or decision criteria, leaving the group procedurally blocked. &
A mild disagreement where the group already agrees on how to compare options. \\
\midrule
\textsc{Uncertainty} &
A needed fact, requirement, or decision criterion is missing, and the ambiguity blocks downstream work. &
A participant simply guesses the answer or makes an external promise that resolves the ambiguity. \\
\midrule
\textsc{Underspecified Plan} &
The group treats a plan as actionable while owners, deadlines, scope, handoffs, or success criteria remain missing. &
The group already notices the missing slots and agrees to fill them before moving forward. \\
\midrule
\textsc{Forgotten Constraint} &
A current proposal violates an earlier constraint, such as a budget cap, deadline, privacy rule, policy, or approval requirement. &
A participant already remembers the constraint and adjusts the plan. \\
\midrule
\textsc{Goal Drift} &
The group moves away from an explicitly stated goal, blocker, or decision before resolving it. &
The group intentionally closes the first goal or explicitly agrees to switch the agenda. \\
\midrule
\textsc{Discussion Loop} &
Participants repeat the same trade-off or pros and cons without adding new information or adopting a decision rule. &
A first-pass comparison where participants are still introducing new evidence. \\
\midrule
\textsc{Participation Imbalance} &
One participant pushes the group toward a decision while affected collaborators or missing perspectives have not been included. &
The quieter participants already state detailed objections, making the issue primarily a conflict. \\
\midrule
\textsc{Risky Commitment} &
The group moves toward an external or high-stakes commitment despite unresolved blockers, missing approval, or unclear fallback plans. &
The group clearly marks the commitment as tentative or conditional on resolving the risk. \\
\bottomrule
\end{tabular}
\caption{Breakdown-specific generation constraints used for synthetic BEAM-derived examples.}
\label{tab:synthetic-breakdown-guide}
\end{table*}

\paragraph{Filtering and labeling.}
Each generated example is checked for three conditions. First, the target breakdown must be visible from the conversation history without relying on hidden labels. Second, the final trigger turn must create a clear decision point where a concise group-facing intervention would be useful. Third, the participants should not already be resolving the issue by themselves. We remove examples where the issue is too implicit, the target label is confounded with another breakdown, or the intervention point is unclear.

\begin{table*}[t]
\centering
\small
\setlength{\tabcolsep}{5pt}
\begin{tabular}{@{}p{0.22\textwidth}p{0.24\textwidth}p{0.47\textwidth}@{}}
\toprule
\textbf{ProACT action / skill} & \textbf{Triggered issue} & \textbf{Purpose} \\
\midrule
\textsc{Silence} &
\textsc{None} &
Stay out of the conversation when participants are already making progress, self-repairing the issue, or when intervention would be duplicative or interruptive. \\
\midrule
Clarification &
\textsc{Uncertainty} &
Ask a focused question when goals, evidence, requirements, or decision criteria are unclear. \\
\midrule
Plan completion &
\textsc{Underspecified Plan} &
Surface missing execution details, such as owner, deadline, scope, handoff, or success criterion. \\
\midrule
Conflict mediation &
\textsc{Conflict} &
Name incompatible goals, constraints, or preferences in a neutral way and help the group compare them. \\
\midrule
Constraint reminder &
\textsc{Forgotten Constraint} &
Remind the group of an earlier constraint, such as a budget, deadline, policy, privacy rule, or approval requirement. \\
\midrule
Goal refocusing &
\textsc{Goal Drift} &
Bring attention back to the stated goal, blocker, or decision that has been left unresolved. \\
\midrule
Loop breaking &
\textsc{Discussion Loop} &
Identify a repeated unresolved trade-off and suggest a lightweight decision rule or next step. \\
\midrule
Participation balancing &
\textsc{Participation Imbalance} &
Invite missing or underrepresented perspectives before the group moves toward a decision. \\
\midrule
Risk check &
\textsc{Risky Commitment} &
Flag unresolved risks, dependencies, missing approvals, or premature commitments before action is taken. \\
\bottomrule
\end{tabular}
\caption{ProACT skill library and action space. The agent either remains silent or selects a targeted collaboration skill based on the detected collaboration issue.}
\label{tab:skill-library}
\label{tab:actions}
\end{table*}

\section{LLM Judge Prompt}
\label{app:judge-prompt}

\definecolor{judgeboxbg}{HTML}{F8FAFC}
\definecolor{judgeboxborder}{HTML}{2563EB}
\definecolor{judgeboxaccent}{HTML}{0F766E}

Table~\ref{tab:judge-prompt} presents the full LLM-as-judge prompt used to evaluate candidate outputs in our benchmark. The prompt asks the judge to assess each action and visible response according to collaborative participation criteria, including appropriateness, non-interruption, conciseness, grounding, neutrality, and intervention quality. We format the prompt as a two-column appendix table to keep the instructions readable.

\begin{table*}[t]
\centering
\begingroup
\setlength{\fboxsep}{10pt}
\setlength{\fboxrule}{0.8pt}
\fcolorbox{promptboxborder}{promptboxbg}{%
\begin{minipage}{0.94\textwidth}
\small
\textbf{\textcolor{promptboxaccent}{Judge prompt used in our benchmark.}}

\medskip
\textbf{System instruction.} You are an expert judge for proactive multi-user collaboration agents. Your job is to evaluate a candidate assistant's response at the current moment in a group conversation. Judge the candidate by human collaborative participation standards, not by whether it is technically smart in isolation. Use criteria from human collaboration and facilitation research: maintaining common ground, supporting shared plans and commitments, managing coordination/articulation work, respecting turn-taking and timing, preserving neutrality and fairness, balancing participation, grounding claims in the visible conversation, being brief, and preserving human agency.

\medskip
A good proactive response should help move the group conversation forward now with a concrete collaborative next step; improve common ground, coordination, plan completeness, risk awareness, or participation balance; avoid interrupting productive human flow or duplicating what humans are already doing; avoid taking sides, taking over the work, or giving an overlong answer; be grounded in the visible conversation; and be concise enough to fit naturally as one group-chat contribution. Silence can be good when speaking would be unnecessary, duplicative, speculative, or interruptive. Judge whether the candidate's action itself is collaboratively appropriate: either a useful, well-timed contribution or appropriate silence. Return JSON only.

\medskip
\textbf{User payload template.} For each evaluated model output, the user message provides the conversation, the candidate response, and the inferred candidate action (\texttt{intervene} if the response is non-empty, otherwise \texttt{silence}). The judge is asked to return: \texttt{candidate\_action}, \texttt{should\_have\_spoken}, \texttt{collaboratively\_appropriate}, \texttt{helps\_move\_conversation\_forward}, \texttt{non\_interruptive}, \texttt{grounded\_in\_context}, \texttt{concise}, \texttt{not\_overbearing}, \texttt{good\_proactive\_response}, \texttt{quality\_score\_1\_to\_5}, and a brief \texttt{rationale}.

\medskip
\textbf{Key criteria.} \texttt{collaboratively\_appropriate} is true when the candidate action is the right collaboration move at that moment: useful well-timed speech, or appropriate silence when speaking would be unnecessary, duplicative, speculative, or interruptive. \texttt{non\_interruptive} is true when the action fits naturally at the current turn without interrupting productive human flow, duplicating a human contribution, or prematurely answering a question addressed to someone else. The 1--5 quality score follows the same rubric: 1 is harmful or disruptive; 3 is mixed but partially useful; 5 is excellent human-like collaborative participation that is clearly needed, minimally intrusive, grounded, concise, and agency-preserving.
\end{minipage}%
}
\endgroup
\caption{LLM-as-judge prompt used to evaluate candidate actions and responses in proactive collaboration.}
\label{tab:judge-prompt}
\end{table*}

\section{ProACT Prompt and Components}
\label{app:proact-prompt-components}

The ProACT agent is implemented with a lightweight skill harness. At inference time, the model receives a collaboration prompt and an explicit list of available ProACT skills, including an applicability check, a breakdown-diagnosis router, targeted collaboration repair skills, an intervention filter, and a final output contract. Table~\ref{tab:proact-agent-prompt} shows the agent-facing prompt template. 

\section{The Use of Large Language Models}  
For this paper, we leveraged GPT-5.4\footnote{\url{https://openai.com/}} and Codex\footnote{\url{https://openai.com/codex/}} to support grammar refinement, LaTeX formatting, and the preparation of figure generation code.  All technical ideas, experimental designs, analyses, conclusions, and writing were
developed and carried out entirely by the authors. The authors have full responsibility for the final
text.  

\begin{table*}[t]
\centering
\begingroup
\setlength{\fboxsep}{10pt}
\setlength{\fboxrule}{0.8pt}
\providecolor{promptboxbg}{HTML}{F9FAFB}
\providecolor{promptboxborder}{HTML}{D1D5DB}
\providecolor{promptboxaccent}{HTML}{111827}
\fcolorbox{promptboxborder}{promptboxbg}{%
\begin{minipage}{0.94\textwidth}
\small
\textbf{\textcolor{promptboxaccent}{ProACT agent prompt template.}}

\medskip
\textbf{System instruction.}
You are a proactive collaboration assistant for multi-user group conversations. Your role is to help the group only when a short intervention can improve collaboration. You should observe the visible conversation, identify whether the current point contains a collaboration breakdown, and decide whether to stay silent or produce a concise group-facing message.

\medskip
\textbf{Input.}
The input is a speaker-attributed conversation history. Each speaker tag denotes a different participant in the collaboration.

\medskip
\textbf{Available ProACT components.}
You may use the ProACT components for applicability checking, breakdown diagnosis, skill routing, collaboration repair, intervention filtering, and final response formatting.

\medskip
\textbf{Decision rule.}
Stay silent when participants are already making progress, self-repairing the issue, or when an assistant response would be duplicative or interruptive. Intervene only when the conversation shows an unresolved collaboration issue and a short group-facing message would likely improve coordination.

\medskip
\textbf{Visible output contract.}
If the decision is silence, produce no visible group-chat text. If the decision is intervention, output only the concise message to the group. Do not include diagnostic labels, skill names, rationales, or meta-commentary in the visible response.
\end{minipage}%
}
\endgroup
\caption{Prompt Template used by ProACT.}
\label{tab:proact-agent-prompt}
\end{table*}

\end{document}